\documentclass[11pt]{article}
\usepackage{acl2016}
\usepackage{times}
\usepackage{latexsym}
\usepackage{bm}
\usepackage{amsfonts}
\usepackage{graphicx}
\usepackage{stmaryrd}
\usepackage{algorithm}
\usepackage{algpseudocode}
\usepackage{multirow}
\usepackage{epstopdf}
\usepackage{caption}

\aclfinalcopy 

\def\argmax{\mathop{\rm argmax}}%

\title{Agreement-based Learning of Parallel Lexicons and Phrases \\ from Non-Parallel Corpora}
\author{
        Chunyang Liu$^\dagger$,
        Yang Liu$^\dagger$$^\#$\thanks{\ \ Corresponding author: Yang Liu.} ,
        Huanbo Luan$^\dagger$,
        Maosong Sun$^\dagger$$^\#$, \and
        Heng Yu$^\ddagger$ \\
        $^\dagger$ State Key Laboratory of Intelligent Technology and Systems \\
                   Tsinghua National Laboratory for Information Science and Technology \\
                   Department of Computer Science and Technology, Tsinghua University, Beijing 100084, China \\
        $^\#$ Jiangsu Collaborative Innovation Center for Language Competence, Jiangsu, China \\
        $^\ddagger$ Samsung R\&D Institute of China, Beijing 100028, China \\
        \small {\tt \{liuchunyang2012,liuyang.china,luanhuanbo\}@gmail.com, sms@tsinghua.edu.cn} \\
        \small {\tt h0517.yu@samsung.com}
       }
\date{}

\begin{document}

\maketitle

\begin{abstract}
We introduce an agreement-based approach to learning parallel lexicons and phrases from non-parallel corpora. The basic idea is to encourage two asymmetric latent-variable translation models (i.e., source-to-target and target-to-source) to agree on identifying latent phrase and word alignments. The agreement is defined at both word and phrase levels. We develop a Viterbi EM algorithm for jointly training the two unidirectional models efficiently. Experiments on the Chinese-English dataset show that agreement-based learning significantly improves both alignment and translation performance.
\end{abstract}

\section{Introduction}
Parallel corpora, which are large collections of parallel texts, serve as an important resource for inducing translation correspondences, either at the level of words \cite{Brown:93,Smadja:94,Wu:94} or phrases \cite{Kupiec:93,Melamed:97,Marcu:02,Koehn:03}. However, the availability of large-scale, wide-coverage corpora still remains a challenge even in the era of big data: parallel corpora are usually only existent for resource-rich languages and restricted to limited domains such as government documents and news articles.

Therefore, intensive attention has been drawn to exploiting non-parallel corpora for acquiring translation correspondences. Most previous efforts have concentrated on learning parallel lexicons from non-parallel corpora, including parallel sentence and lexicon extraction via bootstrapping \cite{Fung:04}, inducing parallel lexicons via canonical correlation analysis \cite{Haghighi:08}, training IBM models on monolingual corpora as decipherment \cite{Ravi:11,Nuhn:12,Dou:14}, and deriving parallel lexicons from bilingual word embeddings \cite{Vulic:13,Mikolov:13,Vulic:15}.

Recently, a number of authors have turned to a more challenging task: learning parallel phrases from non-parallel corpora \cite{Zhang:13,Dong:15}. Zhang and Zong \shortcite{Zhang:13} present a method for retrieving parallel phrases from non-parallel corpora using a seed parallel lexicon. Dong et al. \shortcite{Dong:15} continue this line of research to further introduce an iterative approach to joint learning of parallel lexicons and phrases. They introduce a corpus-level latent-variable translation model in a non-parallel scenario and develop a training algorithm that alternates between (1) using a parallel lexicon to extract parallel phrases from non-parallel corpora and (2) using the extracted parallel phrases to enlarge the parallel lexicon. They show that starting from a small seed lexicon, their approach is capable of learning both new words and phrases gradually over time.

However, due to the structural divergence between natural languages as well as the presence of noisy data, only using asymmetric translation models might be insufficient to accurately identify parallel lexicons and phrases from non-parallel corpora. Dong et al. \shortcite{Dong:15} report that the accuracy on Chinese-English dataset is only around 40\% after running for 70 iterations. In addition, their approach seems prone to be affected by noisy data in non-parallel corpora as the accuracy drops significantly with the increase of noise.

Since asymmetric word alignment and phrase alignment models are usually complementary, it is natural to combine them to make more accurate predictions. In this work, we propose to introduce agreement-based learning \cite{Liang:06,Liang:08} into extracting parallel lexicons and phrases from non-parallel corpora. Based on the latent-variable model proposed by Dong et al. \shortcite{Dong:15}, we propose two kinds of loss functions to take into account the agreement between both phrase alignment and word alignment in two directions. As the inference is intractable, we resort to a Viterbi EM algorithm to train the two models efficiently. Experiments on the Chinese-English dataset show that agreement-based learning is more robust to noisy data and leads to substantial improvements in phrase alignment and machine translation evaluations.

\section{Background}

Given a monolingual corpus of source language phrases $E=\{\mathbf{e}^{(s)}\}_{s=1}^{S}$ and a monolingual corpus of target language phrases $F=\{\mathbf{f}^{(t)}\}_{t=1}^{T}$, we assume there exists a parallel corpus $D=\{\langle \mathbf{e}^{(s)}, \mathbf{f}^{(t)} \rangle | \mathbf{e}^{(s)} \leftrightarrow \mathbf{f}^{(t)} \}$, where $\mathbf{e}^{(s)} \leftrightarrow \mathbf{f}^{(t)}$ denotes that $\mathbf{e}^{(s)}$ and $\mathbf{f}^{(t)}$ are translations of each other.

As a long sentence in $E$ is usually unlikely to have an translation in $F$ and vise versa, most previous efforts build on the assumption that {\em phrases} are more likely to have translational equivalents on the other side \cite{Munteanu:06,Cettolo:10,Zhang:13,Dong:15}. Such a set of phrases can be constructed by collecting either constituents of parsed sentences or strings with hyperlinks on webpages (e.g., Wikipedia). Therefore, we assume the two monolingual corpora are readily available and focus on how to extract $D$ from $E$ and $F$.

To address this problem, Dong et al. \shortcite{Dong:15} introduce a corpus-level latent-variable translation model in a non-parallel scenario:
\begin{eqnarray}
P(F|E; \bm{\theta})= \sum_{\mathbf{m}} \underbrace{ P(F, \mathbf{m}|E; \bm{\theta})}_{\textrm{\em phrase alignment}},
\end{eqnarray}
where $\mathbf{m}$ is {\em phrase alignment} and $\bm{\theta}$ is a set of model parameters. Each target phrase $\mathbf{f}^{(t)}$ is restricted to connect to exactly one source phrase: $\mathbf{m} = (\mathbf{m}_1, \dots, \mathbf{m}_t, \dots \mathbf{m}_T)$, where $\mathbf{m}_t \in \{0, 1,\dots, S\}$. For example, $\mathbf{m}_t = s$ denotes that $\mathbf{f}^{(t)}$ is aligned to $\mathbf{e}^{(s)}$. Note that $\mathbf{e}^{(0)}$ represents an empty source phrase.

They follow IBM Model 1 \cite{Brown:93} to further decompose the model as
\begin{eqnarray}
P(F, \mathbf{m}|E; \bm{\theta}) = \frac{p(T|S)}{(S+1)^T} \prod_{t=1}^{T}P(\mathbf{f}^{(t)}|\mathbf{e}^{(\mathbf{m}_t)}; \bm{\theta}),
\end{eqnarray}
where $P(\mathbf{f}^{(t)}|\mathbf{e}^{(\mathbf{m}_t)}; \bm{\theta})$ is a {\em phrase translation} model that can be further defined as
\begin{eqnarray}
P(\mathbf{f}^{(t)}|\mathbf{e}^{(\mathbf{m}_t)}; \bm{\theta}) \quad \quad \quad \quad \quad \quad \quad \quad \quad \nonumber \\
= \delta(\mathbf{m}_t, 0) \epsilon + \quad \quad \quad \quad \quad \quad \quad \quad \quad \quad \ \ \nonumber \\
 \big(1 - \delta(\mathbf{m}_t, 0)\big)\sum_{\mathbf{a}}\underbrace{P(\mathbf{f}^{(t)}, \mathbf{a}|\mathbf{e}^{(\mathbf{m}_t)}; \bm{\theta})}_{\textrm{\em word alignment}}. \
\end{eqnarray}

\begin{figure*}[!t]
\centering
\includegraphics[width=1.0\textwidth]{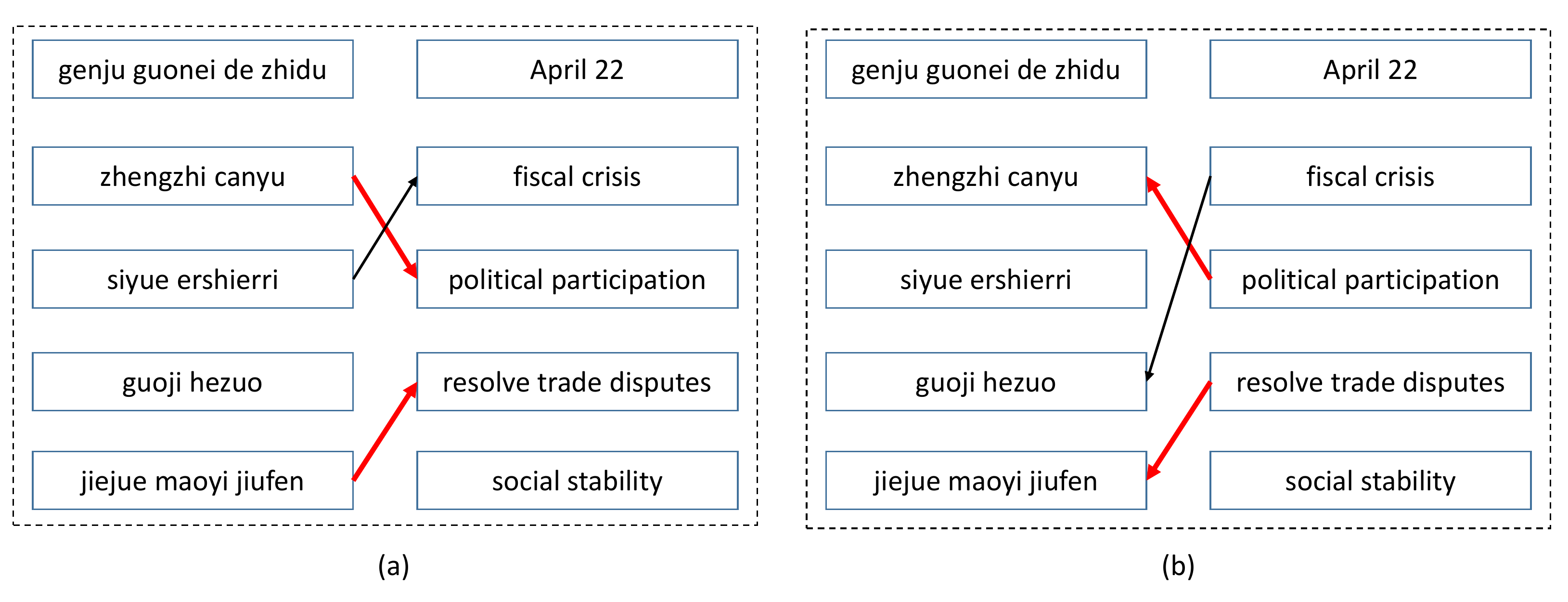}
\caption{Agreement between (a) Chinese-to-English and (b) English-to-Chinese phrase alignments. The arrows indicate translation directions. The links on which two models agree are highlighted in bold red. The {\em outer agreement} loss function (see Eq. (\ref{eq:outer})) aims to encourage the agreement at the phrase level.} \label{fig_phrase_align}
\end{figure*}

Dong et al. \shortcite{Dong:15} distinguish between {\em empty} and {\em non-empty} phrase translations. If a target phrase $\mathbf{f}^{(t)}$ is aligned to the empty source phrase $\mathbf{e}^{(0)}$ (i.e., $\mathbf{m}_t = 0$), they set the phrase translation probability to a fixed number $\epsilon$. Otherwise, conventional word alignment models such as IBM Model 1 can be used for non-empty phrase translation:
\begin{eqnarray}
P(\mathbf{f}^{(t)}, \mathbf{a}|\mathbf{e}^{(\mathbf{m}_t)}; \bm{\theta}) \quad \quad \quad \quad \quad \ \ \nonumber \\
= \frac{p(J^{(t)}|I^{(\mathbf{m}_t)})}{(I^{(\mathbf{m}_t)} + 1)^{J^{(t)}}} \prod_{j=1}^{J^{(t)}}p(\mathbf{f}^{(t)}_j|\mathbf{e}^{(\mathbf{m}_t)}_{\mathbf{a}_j}),
\end{eqnarray}
where $p(J|I)$ is a {\em length} model and $p(f|e)$ is a {\em translation} model. We use $J^{(t)}$ to denote the length of $\mathbf{f}^{(t)}$.

Therefore, the latent-variable model involves two kinds of latent structures: (1) {\em phrase alignment} $\mathbf{m}$ between source and target phrases, (2) {\em word alignment} $\mathbf{a}$ between source and target words within phrases.

Given the two monolingual corpora $E$ and $F$, the training objective is to maximize the likelihood of the training data:
\begin{eqnarray}
\bm{\theta}^{*} &=& \argmax_{\bm{\theta}}\Big\{ \mathcal{L}(\bm{\theta}) \Big\},
\end{eqnarray}
where
\begin{eqnarray}
\mathcal{L}(\bm{\theta}) &=& \log P(F|E; \bm{\theta}) - \nonumber \\
&& \sum_{I} \lambda_I \Big(\sum_{J}p(J|I) - 1 \Big) - \nonumber \\
&& \sum_{e} \gamma_e \Big(\sum_{f} p(f|e) - 1 \Big) - \nonumber \\
&& \sum_{f}\sum_{e} \sigma(f,e,\mathbf{d})\log \frac{\sigma(f,e, \mathbf{d})}{p(f|e)}.
\end{eqnarray}
Note that $\mathbf{d}$ is a small seed parallel lexicon for initializing training \footnote{Due to the difficulty of learning translation correspondences from non-parallel corpora, many authors have assumed that a small seed lexicon is readily available \cite{Gaussier:04,Zhang:13,Vulic:13,Mikolov:13,Dong:15}.} and $\sigma(f,e, \mathbf{d})$ checks whether an entry $\langle f,e \rangle$ exists in $\mathbf{d}$.

Given the monolingual corpora and the optimized model parameters, the Viterbi phrase alignment is calculated as
\begin{eqnarray}
\mathbf{m}^{*} &=& \argmax_{\mathbf{m}}\Big\{ P(F,\mathbf{m}|E;\bm{\theta}^{*}) \Big\} \\
&=& \argmax_{\mathbf{m}} \Bigg\{ \prod_{t=1}^{T} P(\mathbf{f}^{(t)}|\mathbf{e}^{(\mathbf{m}_t)}; \bm{\theta}^{*}) \Bigg\}.
\end{eqnarray}

Finally, parallel lexicons can be derived from the translation probability table of IBM model 1 $\bm{\theta}^{*}$ and parallel phrases can be collected from the Viterbi phrase alignment $\mathbf{m}^{*}$. This process iterates and enlarges parallel lexicons and phrases gradually over time.

As it is very challenging to extract parallel phrases from non-parallel corpora, unidirectional models might only capture partial aspects of translation modeling on non-parallel corpora. Indeed, Dong et al. \shortcite{Dong:15} find that the accuracy of phrase alignment is only around 50\% on the Chinese-English dataset. More importantly, their approach seems to be vulnerable to noise as the accuracy drops significantly with the increase of noise. As source-to-target and target-to-source translation models are usually complementary \cite{Och:03,Koehn:03,Liang:06}, it is appealing to combine them to improve alignment accuracy.

\section{Approach}

\subsection{Agreement-based Learning}

The basic idea of our work is to encourage the source-to-target and target-to-source translation models to agree on both phrase and word alignments.

For example, Figure \ref{fig_phrase_align} shows  two example Chinese-to-English and English-to-Chinese phrase alignments on the same non-parallel data. As each model only captures partial aspects of translation modeling, our intuition is that the links on which two models agree (highlighted in red) are more likely to be correct.

More formally, let $P(F|E; \overrightarrow{\bm{\theta}})$ be a source-to-target translation model and $P(E|F; \overleftarrow{\bm{\theta}})$ be a target-to-source model, where $\overrightarrow{\bm{\theta}}$ and $\overleftarrow{\bm{\theta}}$ are corresponding model parameters. We use $\overrightarrow{\mathbf{m}} = (\overrightarrow{\mathbf{m}}_1, \dots, \overrightarrow{\mathbf{m}}_t, \dots,  \overrightarrow{\mathbf{m}}_T)$ to denote source-to-target phrase alignment. Likewise, the target-to-source phrase alignment is denoted by  $\overleftarrow{\mathbf{m}} = (\overleftarrow{\mathbf{m}}_1, \dots, \overleftarrow{\mathbf{m}}_s, \dots,  \overleftarrow{\mathbf{m}}_S)$.

To ease the comparison between $\overrightarrow{\mathbf{m}}$ and $\overleftarrow{\mathbf{m}}$, we represent them as sets of non-empty links equivalently:
\begin{eqnarray}
\overrightarrow{\mathbf{m}} &=& \Big\{ \langle \overrightarrow{\mathbf{m}}_t, t \rangle |\overrightarrow{\mathbf{m}}_t \ne 0  \Big \} \\
\overleftarrow{\mathbf{m}} &=& \Big\{ \langle s, \overleftarrow{\mathbf{m}}_s \rangle |\overleftarrow{\mathbf{m}}_s \ne 0  \Big \}.
\end{eqnarray}

For example, suppose the source-to-target and target-to-source phrase alignments are $\overrightarrow{\mathbf{m}}=(2, 3, 0, 0)$ and $\overleftarrow{\mathbf{m}}=(0, 1, 2)$. The equivalent link sets are $\overrightarrow{\mathbf{m}} = \{ \langle 2, 1 \rangle,  \langle 3, 2 \rangle   \}$
and $\overleftarrow{\mathbf{m}} = \{ \langle 2, 1 \rangle,  \langle 3, 2 \rangle    \}$. Therefore, $\overrightarrow{\mathbf{m}}$ is said to be {\em equal} to $\overleftarrow{\mathbf{m}}$ (i.e., $\delta(\overrightarrow{\mathbf{m}}, \overleftarrow{\mathbf{m}}) = 1$).

Following Liang et al. \shortcite{Liang:06}, we introduce a new training objective that favors the agreement between two unidirectional models:
\begin{eqnarray}
\mathcal{J}(\overrightarrow{\bm{\theta}}, \overleftarrow{\bm{\theta}}) \quad \quad \quad \quad \quad \quad \quad \quad \quad \quad \quad \quad \quad \nonumber \\
= \log P(F|E; \overrightarrow{\bm{\theta}}) + \log P(E|F; \overleftarrow{\bm{\theta}}) - \quad \ \ \ \ \nonumber \\
\log \sum_{\overrightarrow{\mathbf{m}},\overleftarrow{\mathbf{m}}}P(\overrightarrow{\mathbf{m}}|E,F; \overrightarrow{\bm{\theta}})P(\overleftarrow{\mathbf{m}}|F,E; \overleftarrow{\bm{\theta}}) \ \ \ \nonumber \\
\times \Delta(E, F, \overrightarrow{\mathbf{m}}, \overleftarrow{\mathbf{m}}, \overrightarrow{\bm{\theta}}, \overleftarrow{\bm{\theta}}), \quad \quad \ \ \
\end{eqnarray}
where the posterior probabilities in two directions are defined as
\begin{eqnarray}
P(\overrightarrow{\mathbf{m}}|E,F; \overrightarrow{\bm{\theta}}) = \prod_{t=1}^{T} \frac{P(\mathbf{f}^{(t)}|\mathbf{e}^{(\overrightarrow{\mathbf{m}}_t)}; \overrightarrow{\bm{\theta}})}{\sum_{s=0}^{S}P(\mathbf{f}^{(t)}|\mathbf{e}^{(s)}; \overrightarrow{\bm{\theta}})} \ \\
P(\overleftarrow{\mathbf{m}}|F,E; \overleftarrow{\bm{\theta}}) = \prod_{s=1}^{S} \frac{P(\mathbf{e}^{(s)}|\mathbf{f}^{(\overleftarrow{\mathbf{m}}_s)}; \overleftarrow{\bm{\theta}})}{\sum_{t=0}^{T}P(\mathbf{e}^{(s)}|\mathbf{f}^{(t)}; \overleftarrow{\bm{\theta}})}.
\end{eqnarray}

The {\em loss function} $\Delta(E, F, \overrightarrow{\mathbf{m}}, \overleftarrow{\mathbf{m}}, \overrightarrow{\bm{\theta}}, \overleftarrow{\bm{\theta}})$ measures the disagreement between the two models.

\subsection{Outer Agreement}

\begin{figure}[!t]
\begin{algorithmic}[1]
\Procedure{ViterbiEM}{$E$, $F$, $\mathbf{d}$}
    \State {Initialize $\bm{\Theta}^{(0)}$}
    \ForAll {$k = 1, \dots, K$}
        \State $\hat{\mathbf{m}}^{(k)} \gets$ \Call{Search}{$E, F, \bm{\Theta}^{(k-1)}$}
        \State $\bm{\Theta}^{(k)} \gets$ \Call{Update}{$E, F, \mathbf{d}, \hat{\mathbf{m}}^{(k)}$}
    \EndFor
    \State \textbf{return} $\hat{\mathbf{m}}^{(K)}, \bm{\Theta}^{(K)}$
\EndProcedure
\end{algorithmic}
\caption{A Viterbi EM algorithm for agreement-based learning of parallel lexicons and phrases from non-parallel corpora. $F$ and $E$ are non-parallel corpora, $\mathbf{d}$ is a seed parallel lexicon, $\bm{\Theta}^{(k)}$ is the set of model parameters at the $k$-th iteration, $\hat{\mathbf{m}}^{(k)}$ is the Viterbi phrase alignment on which two models agree at the $k$-th iteration.} \label{fig:viterbi_em}
\end{figure}

\subsubsection{Definition}

 A straightforward loss function is to force the two models to generate identical phrase alignments:
\begin{eqnarray}
\Delta_{\textrm{outer}}(E, F, \overrightarrow{\mathbf{m}}, \overleftarrow{\mathbf{m}}, \overrightarrow{\bm{\theta}}, \overleftarrow{\bm{\theta}}) = 1 - \delta(\overrightarrow{\mathbf{m}}, \overleftarrow{\mathbf{m}}). \label{eq:outer}
\end{eqnarray}

We refer to Eq. (\ref{eq:outer}) as {\em outer agreement} since it only considers phrase alignment and ignores the word alignment within aligned phrases.

\subsubsection{Training Objective}

Since the outer agreement forces two models to generate identical phrase alignments, the training objective can be written as
\begin{eqnarray}
\mathcal{J}_{\textrm{outer}}(\overrightarrow{\bm{\theta}}, \overleftarrow{\bm{\theta}}) \quad \quad \quad \quad \quad \quad \quad \quad \quad \ \ \ \nonumber \\
= \log P(F|E; \overrightarrow{\bm{\theta}}) + \log P(E|F; \overleftarrow{\bm{\theta}}) + \ \ \ \nonumber \\
 \log \sum_{\mathbf{m}}P(\mathbf{m}|E,F;\overrightarrow{\bm{\theta}})P(\mathbf{m}|F,E; \overleftarrow{\bm{\theta}}),
\end{eqnarray}
where $\mathbf{m}$ is a phrase alignment on which two models agree.

The partial derivatives of the training objective with respect to source-to-target model parameters $\overrightarrow{\bm{\theta}}$ are given by
\begin{eqnarray}
\frac{\partial \mathcal{J}_{\textrm{outer}}(\overrightarrow{\bm{\theta}}, \overleftarrow{\bm{\theta}})}{\partial \overrightarrow{\bm{\theta}}} \quad \quad \quad \quad \quad \quad \quad \quad \nonumber \\
= \frac{\partial P(F|E; \overrightarrow{\bm{\theta}}) / \partial \overrightarrow{\bm{\theta}}}{P(F|E; \overrightarrow{\bm{\theta}})}  + \quad \quad \quad \quad \quad \ \ \nonumber \\
 \frac{\mathbb{E}_{\mathbf{m}|F, E; {\small \overleftarrow{ \bm{\theta}}}}\Big[\partial P(F|E; \overrightarrow{\bm{\theta}}) / \partial \overrightarrow{\bm{\theta}}\Big]}{\sum_{\mathbf{m}}P(\mathbf{m}|E,F;\overrightarrow{\bm{\theta}})P(\mathbf{m}|F,E; \overleftarrow{\bm{\theta}})}. \label{eq:outer_derivative}
\end{eqnarray}
The partial derivatives with respect to $\overleftarrow{\bm{\theta}}$ are defined likewise.

\subsubsection{Training Algorithm}

As the expectation in Eq. (\ref{eq:outer_derivative}) is usually intractable to calculate due to the exponential search space of phrase alignment, we follow Dong et al. \shortcite{Dong:15} to use a Viterbi EM algorithm instead.

As shown in Figure \ref{fig:viterbi_em}, the algorithm takes a set of source phrases $E$, a set of target phrases $F$, and a seed parallel lexicon $\mathbf{d}$ as input (line 1). After initializing model parameters $\bm{\Theta}=\{ \overrightarrow{\bm{\theta}}, \overleftarrow{\bm{\theta}} \}$ (line 2), the algorithm calls the procedure \textproc{Align}$(F, E, \bm{\Theta})$ to compute the Viterbi phrase alignment between $E$ and $F$ on which two models agree. Then, the algorithm updates the two models by normalizing counts collected from the Viterbi phrase alignment. The process iterates for $K$ iterations and returns the final  Viterbi phrase alignment and model parameters.

\subsubsection{Computing Viterbi Phrase Alignments}
The procedure \textproc{Align}$(F, E, \bm{\Theta})$ computes the Viterbi phrase alignment $\hat{\mathbf{m}}$ between $E$ and $F$ on which two models agree as follows:
\begin{eqnarray}
\hat{\mathbf{m}} = \argmax_{\mathbf{m}}\Big\{ P(\mathbf{m}|E, F; \overrightarrow{\bm{\theta}})P(\mathbf{m}|F, E; \overleftarrow{\bm{\theta}}) \Big\}.
\end{eqnarray}

Unfortunately, due to the exponential search space of phrase alignment, computing $\hat{\mathbf{m}}$ is also intractable. As a result, we approximate it as the intersection of two unidirectional Viterbi phrase alignments:
\begin{eqnarray}
\hat{\mathbf{m}} \approx \overrightarrow{\mathbf{m}}^{*} \cap \overleftarrow{\mathbf{m}}^{*}, \label{eq:intersection}
\end{eqnarray}
where the unidirectional Viterbi phrase alignments are calculated as
\begin{eqnarray}
\overrightarrow{\mathbf{m}}^{*} = \argmax_{\overrightarrow{\mathbf{m}}}\Bigg\{ \prod_{t=1}^{T} P(\mathbf{f}^{(t)}|\mathbf{e}^{(\overrightarrow{\mathbf{m}}_t)}; \overrightarrow{\bm{\theta}}) \Bigg\} \ \ \label{eq:calouter}\\
\overleftarrow{\mathbf{m}}^{*} = \argmax_{\overleftarrow{\mathbf{m}}}\Bigg\{ \prod_{s=1}^{S} P(\mathbf{e}^{(s)}|\mathbf{f}^{(\overleftarrow{\mathbf{m}}_s)}; \overleftarrow{\bm{\theta}}) \Bigg\}.
\end{eqnarray}

The source-to-target Viterbi phrase alignment is calculated as
\begin{eqnarray}
\overrightarrow{\mathbf{m}}^{*} = \argmax_{\overrightarrow{\mathbf{m}}}\Big\{ P(\mathbf{\overrightarrow{\mathbf{m}}}|E, F; \overrightarrow{\bm{\theta}}) \Big\} \quad \quad \ \\
= \argmax_{\overrightarrow{\mathbf{m}}}\Big\{ \prod_{t=1}^{T} P(\mathbf{f}^{(t)}|\mathbf{e}^{({\tiny \overrightarrow{\mathbf{m}}}_t)}; \overrightarrow{\bm{\theta}}) \Big\}. \label{eq:two_parts}
\end{eqnarray}

Dong et al. \shortcite{Dong:15} indicate that computing the Viterbi alignment for individual target phrases is {\em independent} and only need to focus on finding the most probable source phrase for each target phrase:
\begin{eqnarray}
\overrightarrow{\mathbf{m}}_t^{*} = \argmax_{s \in \{0,1,\dots,S\}}\Big\{ P(\mathbf{f}^{(t)}|\mathbf{e}^{(s)}; \overrightarrow{\bm{\theta}}) \Big\}.
\end{eqnarray}

This can be cast as a translation retrieval problem \cite{Zhang:13,Dong:14}. Please refer to \cite{Dong:15} for more details. The target-to-source Viterbi phrase alignment can be calculated similarly.

\subsubsection{Updating Model Parameters}

Following Liang et al. \shortcite{Liang:06}, we collect counts of model parameters only from the agreement term.\footnote{We experimented with collecting counts from both the unidirectional and agreement terms but obtained much worse results than counting only from the agreement term.}

Given the agreed Viterbi phrase alignment $\hat{\mathbf{m}}$, the count of the source-to-target length model $p(J|I)$ is given by
\begin{eqnarray}
c(J|I;E, F)  = \sum_{\langle s, t \rangle \in \hat{\mathbf{m}}} \delta(J^{(t)}, J) \delta(I^{(s)}, I).   \label{eq:length_count}
\end{eqnarray}

The new length probabilities can be obtained by
\begin{eqnarray}
p(J|I) = \frac{c(J|I; E, F)}{\sum_{J'} c(J'|I;E, F)}.
\end{eqnarray}

The count of the source-to-target translation model $p(f|e)$ is given by
\begin{eqnarray}
&& c(f|e; E, F) \nonumber \\
&=& \sum_{\langle s, t \rangle \in \hat{\mathbf{m}}} \frac{p(f|e) }{\sum_{i=0}^{I^{(s)}}p(f|\mathbf{e}^{(s)}_i)}  \times \nonumber \\
&& \ \ \ \ \ \ \ \ \ \ \ \sum_{j=1}^{J^{(t)}}\delta(f, \mathbf{f}^{(t)}_j) \sum_{i=0}^{I^{(s)}}\delta(e, \mathbf{e}^{(s)}_i)   \nonumber \\
&& + \sigma(f, e, \mathbf{d}).
\end{eqnarray}

The new translation probabilities can be obtained by
\begin{eqnarray}
p(f|e) = \frac{c(f|e; E, F)}{\sum_{f'} c(f'|e;E, F)}.
\end{eqnarray}

Counts of target-to-source length and translation models can be calculated in a similar way.

\subsection{Inner Agreement}

\begin{figure}[!t]
\centering
\includegraphics[width=0.47\textwidth]{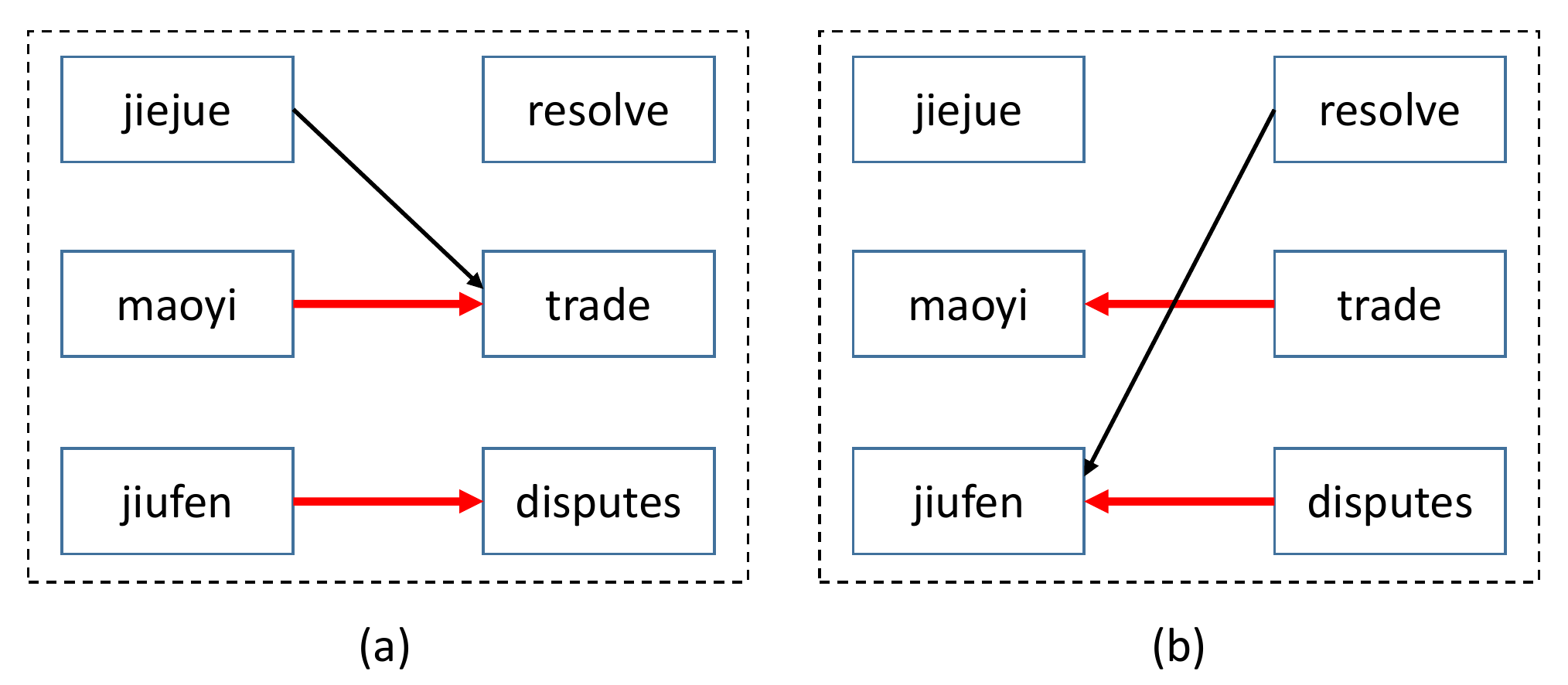}
\caption{Agreement between (a) Chinese-to-English and (b) English-to-Chinese word alignments. The links on which two models agree are highlighted in red. The {\em inner agreement} loss function (see Eq. (\ref{eq:inner})) aims to encourage the agreement at both the phrase and word levels.} \label{fig_example_inner}
\end{figure}

\subsubsection{Definition}
As the outer agreement only considers the phrase alignment, the {\em inner agreement} takes both phrase alignment and word alignment into consideration:
\begin{eqnarray}
\Delta_{\textrm{inner}}(E, F, \overrightarrow{\mathbf{m}}, \overleftarrow{\mathbf{m}}, \overrightarrow{\bm{\theta}}, \overleftarrow{\bm{\theta}}) \quad \quad \quad \quad \nonumber \\
= -\delta(\overrightarrow{\mathbf{m}}, \overleftarrow{\mathbf{m}}) \times \quad \quad \quad \quad \quad \quad \quad \quad \quad \ \ \ \nonumber \\
 \sum_{\langle s,t \rangle \in \overrightarrow{\mathbf{m}}} \sum_{\overrightarrow{\mathbf{a}},\overleftarrow{\mathbf{a}}}P(\overrightarrow{\mathbf{a}}|\mathbf{e}^{(s)}, \mathbf{f}^{(t)}; \overrightarrow{\bm{\theta}}) \times \ \nonumber \\
 P(\overleftarrow{\mathbf{a}}|\mathbf{f}^{(t)}, \mathbf{e}^{(s)}; \overleftarrow{\bm{\theta}}) \times \ \nonumber \\
 \delta(\overrightarrow{\mathbf{a}}, \overleftarrow{\mathbf{a}}). \quad \quad \quad \quad \quad  \label{eq:inner}
\end{eqnarray}

For example, Figure \ref{fig_example_inner} shows two examples of Chinese-to-English and English-to-Chinese word alignments. The shared links are highlighted in red. Our intuition is that a source phrase and a target phrase are more likely to be translations of each other if the two translation models also agree on word alignment within aligned phrases.

\subsubsection{Training Objective and Algorithm}
The training objective for inner agreement is given by
\begin{eqnarray}
&&\mathcal{J}_{\textrm{inner}}(\overrightarrow{\bm{\theta}}, \overleftarrow{\bm{\theta}})   \nonumber \\
&=& \log P(F|E; \overrightarrow{\bm{\theta}}) + \log P(E|F; \overleftarrow{\bm{\theta}}) +  \nonumber \\
&& \log \sum_{\mathbf{m}} P(\mathbf{m}|E, F; \overrightarrow{\bm{\theta}}) P(\mathbf{m}|F, E; \overleftarrow{\bm{\theta}}) \times  \nonumber \\
&& \ \ \ \ \ \ \ \ \ \sum_{\langle s,t \rangle \in \mathbf{m}} \sum_{\mathbf{a}} P(\mathbf{a}| \mathbf{e}^{(s)}, \mathbf{f}^{(t)}; \overrightarrow{\bm{\theta}}) \times \nonumber \\
&& \quad \quad \quad \quad \quad \ \ \ \ \ \ \ P(\mathbf{a}|  \mathbf{f}^{(t)}, \mathbf{e}^{(s)}; \overleftarrow{\bm{\theta}}).
\end{eqnarray}

We still use the Viterbi EM algorithm as shown in Figure \ref{fig:viterbi_em} for training the two models.

\subsubsection{Computing Viterbi Phrase Alignments}

The agreed Viterbi phrase alignment is defined as
\begin{eqnarray}
\hat{\mathbf{m}} = \argmax_{\mathbf{m}}\Big\{ P(\mathbf{m}|E, F; \overrightarrow{\bm{\theta}}) P(\mathbf{m}|F, E; \overleftarrow{\bm{\theta}})  \nonumber \\
 \times \sum_{\langle s,t \rangle \in \mathbf{m}} \sum_{\mathbf{a}} P(\mathbf{a}| \mathbf{e}^{(s)}, \mathbf{f}^{(t)}; \overrightarrow{\bm{\theta}})  \nonumber \\
 \times P(\mathbf{a}|  \mathbf{f}^{(t)}, \mathbf{e}^{(s)}; \overleftarrow{\bm{\theta}})  \Big\}.
\end{eqnarray}

As computing $\hat{\mathbf{m}}$ is intractable, we still approximate it using the intersection of two unidirectional Viterbi phrase alignments (see Eq. (\ref{eq:intersection})). The source-to-target Viterbi phrase alignment is calculated as
\begin{eqnarray}
\overrightarrow{\mathbf{m}}^{*} = \argmax_{\overrightarrow{\mathbf{m}}}\Big\{ P(\overrightarrow{\mathbf{m}}|E, F; \overrightarrow{\bm{\theta}}) \times \quad \quad \quad \quad \ \nonumber \\
\sum_{\langle s,t \rangle \in {\tiny \overrightarrow{\mathbf{m}}}} \sum_{j=1}^{J^{(t)}}\sum_{i=1}^{I^{(s)}}P(\langle i, j \rangle| \mathbf{e}^{(s)}, \mathbf{f}^{(t)}; \overrightarrow{\bm{\theta}}) \times \nonumber \\
P(\langle i, j \rangle| \mathbf{f}^{(t)}, \mathbf{e}^{(s)}; \overleftarrow{\bm{\theta}}) \Big\}, \ \ \label{eq:inner_viterbi}
\end{eqnarray}
where $P(\langle i, j \rangle| \mathbf{e}^{(s)}, \mathbf{f}^{(t)}; \overrightarrow{\bm{\theta}})$ is source-to-target link posterior probability of the link $\langle i, j\rangle$ being present (or absent) in the word alignment according to the source-to-target model, $P(\langle i, j \rangle| \mathbf{f}^{(t)}, \mathbf{e}^{(s)}; \overleftarrow{\bm{\theta}})$ is target-to-source link posterior probability. We follow Liang et al. \shortcite{Liang:06} to use the product of link posteriors to encourage the agreement at the level of word alignment.

We use a coarse-to-fine approach \cite{Dong:15} to compute the Viterbi alignment: first retrieving a coarse set of candidate source phrases using translation probabilities and then selecting the candidate with the highest score according to Eq. (\ref{eq:inner_viterbi}). The target-to-source Viterbi phrase alignment can be calculated similarly.

\subsubsection{Updating Model Parameters}
Given the agreed Viterbi phrase alignment $\hat{\mathbf{m}}$, the count of the source-to-target length model $p(J|I)$ is still given by Eq. (\ref{eq:length_count}). The count of the translation model $p(f|e)$ is calculated as
\begin{eqnarray}
&& c(f|e; E, F)  \nonumber \\
&=& \sum_{\langle s, t \rangle \in \hat{\mathbf{m}}} \sum_{i=1}^{I^{(s)}}\sum_{j=1}^{J^{(t)}}P(\langle i, j \rangle| \mathbf{e}^{(s)}, \mathbf{f}^{(t)}; \overrightarrow{\bm{\theta}}) \times \nonumber \\
&& \quad \quad \quad \quad \quad \quad   P(\langle i, j \rangle| \mathbf{f}^{(t)}, \mathbf{e}^{(s)}; \overleftarrow{\bm{\theta}}) \times \nonumber \\
&& \quad \quad \quad \quad \quad \quad    \delta(f, \mathbf{f}^{(t)}) \delta(e, \mathbf{e}^{(s)})  \nonumber \\
&& + \sigma(f, e, \mathbf{d}).
\end{eqnarray}

Counts of target-to-source length and translation models can be calculated in a similar way.

\section{Experiments}

In this section, we evaluate our approach in two tasks: phrase alignment (Section 4.1) and machine translation (Section 4.2).

\subsection{Alignment Evaluation}
\subsubsection{Evaluation Metrics}
Given two monolingual corpora $E$ and $F$, we suppose there exists a ground truth parallel corpus $G$ and denote an extracted parallel corpus as $D$. The quality of an extracted parallel corpus can be measured by $\textrm{F1} = 2|D \cap G| / (|D| + |G|)$.

\subsubsection{Data Preparation}

Although it is appealing to apply our approach to dealing with real-world non-parallel corpora, it is time-consuming and labor-intensive to manually construct a ground truth parallel corpus. Therefore, we follow Dong et al. \shortcite{Dong:15} to build synthetic $E$, $F$, and $G$ to facilitate the evaluation.

We first extract a set of parallel phrases from a sentence-level parallel corpus using the state-of-the-art phrase-based translation system Moses \cite{Koehn:07} and discard low-probability parallel phrases. Then, $E$ and $F$ can be constructed by corrupting the parallel phrase set by adding irrelevant source and target phrases randomly. Note that the parallel phrase set can serve as the ground truth parallel corpus $G$. We refer to the non-parallel phrases in $E$ and $F$ as {\em noise}.

From LDC Chinese-English parallel corpora, we constructed a {\em development set} and a {\em test set}. The development set contains 20K parallel phrases, 20K noisy Chinese phrases, and 20K noisy English phrases. The test test contains 20K parallel phrases, 180K noisy Chinese phrases, and 180K noisy English phrases. The seed parallel lexicon contains 1K entries.

\begin{figure}[!t]
\centering
\includegraphics[width=0.53\textwidth]{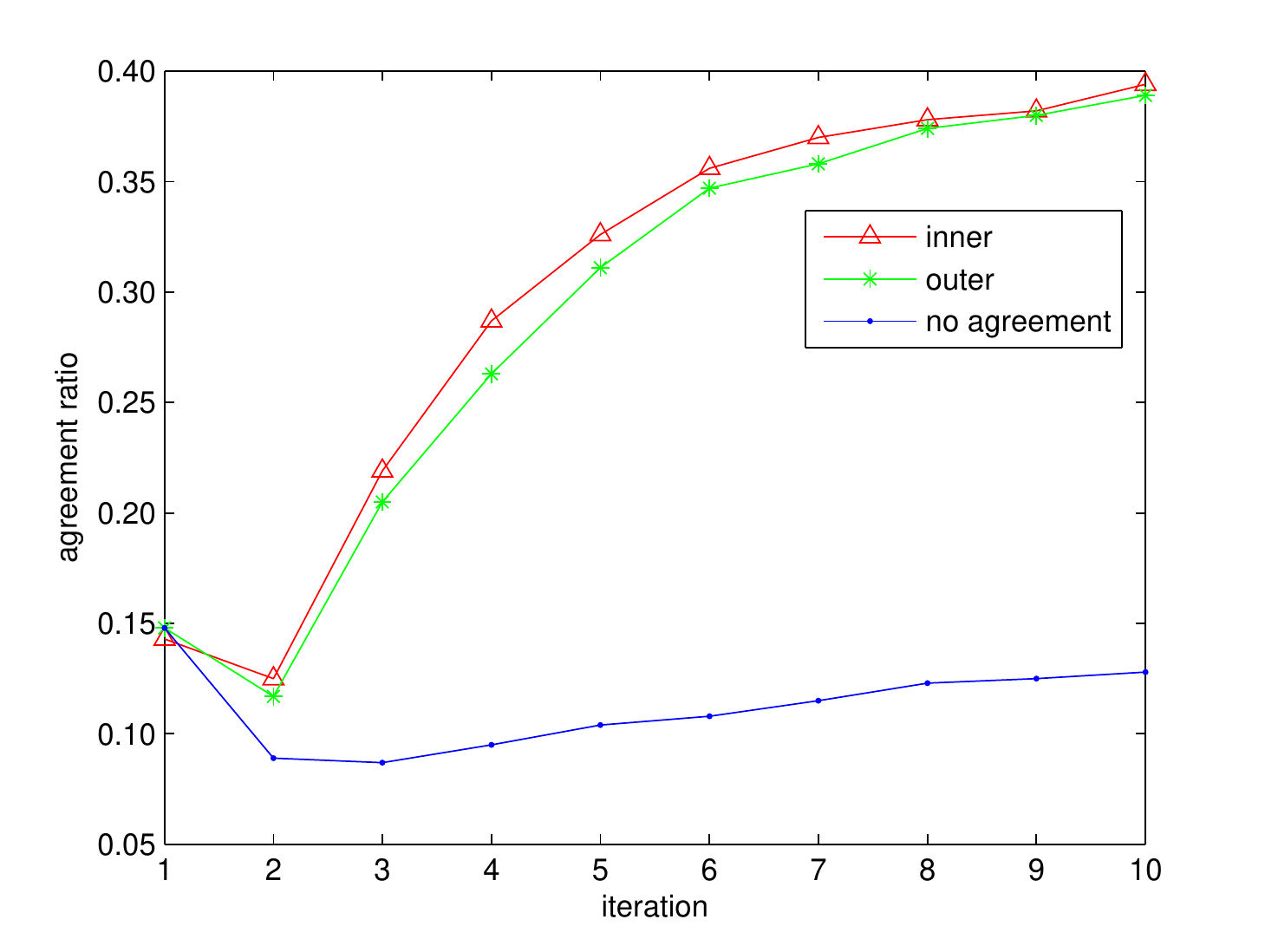}
\caption{Comparison of agreement ratios on the development set.} \label{fig_ratio}
\end{figure}

\begin{table}[!t]
\centering
\begin{tabular}{|r||r|r|r|r|}
\hline
seed & C $\rightarrow$ E & E $\rightarrow$ C & Outer & Inner \\
\hline
50 & 4.1 & 4.8 & 60.8 & 66.2 \\
100 & 5.1 & 5.5 & 65.6 & 69.8 \\
500 & 7.5 & 8.4 & 70.4 & 72.5 \\
1,000 & 22.4 & 23.1 & 73.6 & 74.3\\
\hline
\end{tabular}
\caption{Effect of seed lexicon size in terms of F1 on the development set.} \label{table:seed}
\end{table}

\subsubsection{Comparison of Agreement Ratios}
We introduce {\em agreement ratio} to measure to what extent two unidirectional models agree on phrase alignment:
\begin{eqnarray}
\textrm{ratio} = \frac{2|\overrightarrow{\mathbf{m}}^{*} \cap \overleftarrow{\mathbf{m}}^{*}|}{|\overrightarrow{\mathbf{m}}^{*}| + |\overleftarrow{\mathbf{m}}^{*}|}.
\end{eqnarray}

Figure \ref{fig_ratio} shows the agreement ratios of independent training (``no agreement''), joint training with the outer agreement (``outer''), and joint training with the inner agreement (``inner''). We find that independently trained unidirectional models hardly agree on phrase alignment, suggesting that each model can only capture partial aspects of translation modeling on non-parallel corpora. In contrast, imposing the agreement term significantly increases the agreement ratios: after 10 iterations, about 40\% of phrase alignment links are shared by two models.

\subsubsection{Effect of Seed Lexicon Size}
Table \ref{table:seed} shows the F1 scores of the Chinese-to-English model (``C $\rightarrow$ E''), the English-to-Chinese model (``E $\rightarrow$ C''), joint learning based on the outer agreement (``outer''), and jointing learning based on the inner agreement (``inner'') over various sizes of seed lexicons on the development set.

We find that agreement-based learning obtains substantial improvements over independent learning across all sizes. More importantly, even with a seed lexicon containing only 50 entries, agreement-based learning is able to achieve F1 scores above 60\%. The inner agreement performs better than the outer agreement by taking the consensus at the word level into account.

\begin{table}[!t]
\centering
\begin{tabular}{|c|c||c|c|c|c|}
\hline
\multicolumn{2}{|c||}{noise} & \multirow{2}{*}{C $\rightarrow$ E} & \multirow{2}{*}{E $\rightarrow$ C} & \multirow{2}{*}{Outer} & \multirow{2}{*}{Inner}  \\
\cline{1-2}
C & E & & & & \\
\hline \hline
0 & 0 & 58.5 & 61.2 & 86.5 & 86.1 \\
0 & 10K & 41.0 & 54.4 & 83.6 & 83.8 \\
0 & 20K & 28.3 & 48.3 & 80.1 & 81.2 \\
10K & 0 & 54.7 & 43.1 & 84.9 & 84.3 \\
20K & 0 & 50.4 & 31.4 & 83.8 & 83.6 \\
10K & 10K & 34.9 & 34.4 & 80.0 & 79.7 \\
20K & 20K & 22.4 & 23.1 & 73.6 & 74.3 \\
\hline
\end{tabular}
\caption{Effect of noise in terms of F1 on the development set.} \label{fig_noises}
\end{table}

\subsubsection{Effect of Noise}

Table \ref{fig_noises} demonstrates the effect of noise on the development set. In row 1, ``0+0'' denotes there is no noise, which can be seen as an upper bound. Adding noise, either on the Chinese side or on the English side, deteriorates the F1 scores for all methods. Adding noise on the English side makes predicting phrase alignment in the C $\rightarrow$ E direction more challenging due to the enlarged search space. The situation is similar in the reverse direction. It is clear that agreement-based learning is more robust to noise: while independent training suffers from a reduction of 40\% in terms of F1 for the ``20K + 20K'' setting, agreement-based learning still achieves F1 scores over 70\%.

\subsubsection{Results}
\begin{figure}[!t]
\centering
\includegraphics[width=0.5\textwidth]{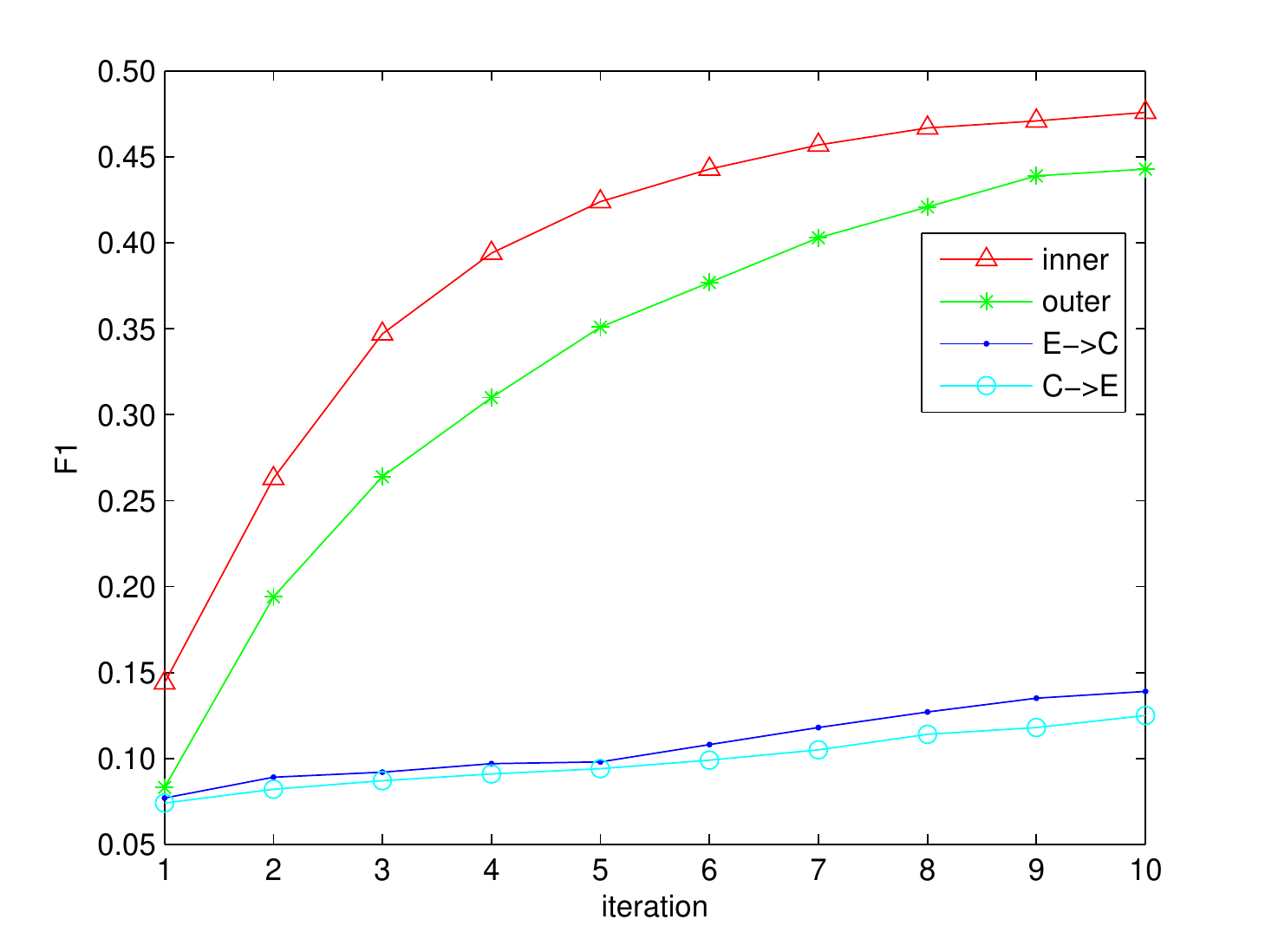}
\caption{Comparison of F1 scores on the test set.} \label{fig_alignment_result}
\end{figure}
Figure \ref{fig_alignment_result} gives the final results on the test set. We find that agreement-based training achieves significant improvements over independent training. By considering the consensus on both phrase and word alignments, the inner agreement significantly outperforms the outer agreement. Notice that \newcite{Dong:15} only add noise on one side while we add noisy phrases on both sides, which makes phrase alignment more challenging.

\begin{table}[!t]
\centering
\begin{tabular}{|c|l|}
\hline
Chinese & jingji \\
\hline
English & {\em economy} \\
\hline \hline
Chinese & {\em jialebi} \\
\hline
English & {\em caribbean}\\
\hline \hline
Chinese & zhengzhi huanjing \\
\hline
English & political environment \\
\hline \hline
Chinese & {\em jiaoyisuo} shichang {\em jiage} zhishu \\
\hline
English & {\em exchange} market {\em price} index \\
\hline \hline
Chinese & {\em qianding} {\em bianjing} {\em maoyi} {\em xieding} \\
\hline
English & {\em signed} {\em border} {\em trade} {\em agreements} \\
\hline
\end{tabular}
\caption{Example learned parallel lexicons and phrases. New words that are not included in the seed lexicon are highlighted in italic.} \label{table:sample}
\end{table}

Table \ref{table:sample} shows example learned parallel words and phrases. The lexicon is built from the translation table by retaining high-probability word pairs. Therefore, our approach is capable of learning both new words and new phrases unseen in the seed lexicon.

\subsection{Translation Evaluation}

\begin{table*}[!t]
\centering
\begin{tabular}{|c||c|c|c|c||c|c|c|c|}
\hline
\multirow{2}{*}{Iteration}  & \multicolumn{4}{c||}{Corpus Size} & \multicolumn{4}{c|}{BLEU} \\
\cline{2-9}
 & E$\rightarrow$ C & C$\rightarrow$ E & Outer & Inner& E$\rightarrow$ C & C$\rightarrow$ E & Outer & Inner\\
\hline \hline
0&\multicolumn{4}{c||}{10k} & \multicolumn{4}{c|}{5.61}\\
\cline{2-9}
1&145k	&162k	&59k	&73k  & 8.65	& 8.90	& 13.53	& 13.74\\
2&195k	&215k	&69k	&101k & 8.82	& 9.47	& 15.26	& 15.61\\
3&209k	&231k	&88k	&132k & 8.42	& 9.29	& 16.88	& 16.94\\
4&214k	&238k	&106k	&159k & 8.46	& 9.27	& 17.15	& 17.83\\
5&217k	&241k	&123k	&181k & 8.87	& 9.40	& 17.94	& 18.89\\
6&219k	&243k	&137k	&197k & 8.52	& 9.30	& 18.56	& 19.47\\
7&222k	&247k	&140k	&207k & 8.81	& 9.22	& 18.72	& 19.46\\
8&224k	&249k	&153k	&220k & 8.71	& 9.26	& 18.84	& 19.50\\
9&227k	&251k	&159k	&233k & 8.92	& 9.35	& 19.05	& 19.63\\
10&229k	&254k	&163k	&239k & 8.33	& 9.06	& 19.39	& 19.78\\
\hline
\end{tabular}
\caption{Results on domain adaptation for machine translation.} \label{table_translation}
\end{table*}

Following \newcite{Zhang:13} and \newcite{Dong:15}, we evaluate our approach on domain adaptation for machine translation.

The data set consists of two in-domain non-parallel corpora and an out-domain parallel corpus. The in-domain non-parallel corpora consists of 2.65M Chinese phrases and 3.67M English phrases extracted from LDC news articles. We use a small out-domain parallel corpus extracted from financial news of FTChina which contains 10K phrase pairs. The task is to extract a parallel corpus from in-domain non-parallel corpora starting from a small out-domain parallel corpus.

We use the state-of-the-art translation system Moses \cite{Koehn:07} and evaluate the performance on Chinese-English NIST datasets. The development set is NIST 2006 and the test set is NIST 2005. The evaluation metric is case-insensitive BLEU4 \cite{Papineni:02}. We use the SRILM toolkit \cite{Stolcke:02} to train a 4-gram English language model on a monolingual corpus with 399M English words.

Table \ref{table_translation} shows the results. At iteration 0, only the out-domain corpus is used and the BLEU score is 5.61. All methods iteratively extract parallel phrases from non-parallel corpora and enlarge the extracted parallel corpus. We find that agreement-based learning achieves much higher BLEU scores while obtains a smaller parallel corpus as compared with independent learning. One possible reason is that the agreement-based learning rules out most unlikely phrase pairs by encouraging consensus between two models.



\section{Conclusion}

We have presented agreement-based training for learning parallel lexicons and phrases from non-parallel corpora. By modeling the agreement on both phrase alignment and word alignment, our approach achieves significant improvements in both alignment and translation evaluations.

In the future, we plan to apply our approach to real-world non-parallel corpora to further verify its effectiveness. It is also interesting to extend the phrase translation model to more sophisticated models such as IBM models 2-5 \cite{Brown:93} and HMM \cite{Vogel:96}.

\section*{Acknowledgments}

We sincerely thank the reviewers for their valuable suggestions. We also thank Meng Zhang, Yankai Lin, Shiqi Shen and Meiping Dong for their insightful discussions. Yang Liu is supported by the National Natural Science Foundation of China (No. 61522204), the 863 Program (2015AA011808), and Samsung R\&D Institute
of China. Huanbo Luan is supported by the National Natural Science Foundation of China (No. 61303075). Maosong Sun is supported by the Major Project of the National Social Science Foundation of China (13\&ZD190).

\bibliography{acl2016_agree}
\bibliographystyle{acl2016}

\end{document}